# A Hybrid Traffic Speed Forecasting Approach Integrating Wavelet Transform and Motif-based Graph Convolutional Recurrent Neural Network

Na Zhang[1], Xuefeng Guan[2*], Jun Cao[3], Xinglei Wang[4], Huayi Wu[5]
[1,2] State Key Laboratory of Information Engineering in Survey, Mapping and Remote Sensing,
zngis@whu.edu.cn, guanxuefeng@whu.edu.cn

## Abstract

Traffic forecasting is crucial for urban traffic management and guidance. However, existing methods rarely exploit the time-frequency properties of traffic speed observations, and often neglect the propagation of traffic flows from upstream to downstream road segments. In this paper, we propose a hybrid approach that learns the spatio-temporal dependency in traffic flows and predicts short-term traffic speeds on a road network. Specifically, we employ wavelet transform to decompose raw traffic data into several components with different frequency sub-bands. A Motif-based Graph Convolutional Recurrent Neural Network (Motif-GCRNN) and Auto-Regressive Moving Average (ARMA) are used to train and predict low-frequency components and high-frequency components, respectively. In the Motif-GCRNN framework, we integrate Graph Convolutional Networks (GCNs) with local sub-graph structures – Motifs – to capture the spatial correlations among road segments, and apply Long Short-Term Memory (LSTM) to extract the short-term and periodic patterns in traffic speeds. Experiments on a traffic dataset collected in Chengdu, China, demonstrate that the proposed hybrid method outperforms six state-of-art prediction methods.

## 1 Introduction

Predicting large-scale traffic conditions in urban areas is of paramount importance in traffic management and travel planning, and a key part of Intelligent Transportation Systems [Mori *et al.*, 2015]. In this paper, we aim to predict short-term traffic speeds in a road network using historical traffic speed observations. Such network-wide traffic forecasting is a challenging task due to the complexity of spatial and temporary dependencies.

The traffic condition on one road segment is affected by its locally adjacent road segments, and nearby segments have a stronger influence on a given segment than more distant segments. In addition, the influence between two adjacent roads is directional: traffic flow on an upstream road spreads to downstream roads and congestion on a downstream road can cause vehicles to decelerate on upstream roads, resulting in delayed congestion on these upstream roads.

Traffic speed time series data exhibits a strong daily periodicity due to human travelling routines (e.g., rush hours when people go to work at morning and go home at evening), which can be referred to as a *daily period* dependence. Furthermore, the traffic speed during one certain time interval is dependent on traffic conditions on that road segment and its neighbors at a recent time, which can be referred to as a *recent trend* dependence.

Learning the complex spatial-temporal dependencies between road segments captures the inherent properties of traffic speeds, that enables accurate forecasting.

In existing studies, there are two kinds of approaches for traffic prediction: time-series analysis based on a statistical approach and data-driven methods based on machine learning. Statistical approaches such as Linear Regression [Nikovski *et al.*, 2005], Kalman Filtering [Chien Steven *et al.*, 2003] and Auto-Regressive Integrated Moving Average (ARIMA) [Lippi *et al.*, 2013] have been widely used in traffic prediction and achieve promising results. However, these models rely on stationary assumptions and have a limited ability to capture complex nonlinear correlations in traffic data. With the increase of traffic datasets and the development of computational power, many research focus on machine learning approaches, such as K-Nearest Neighbors (KNN) [Rice *et al.*, 2004], Support Vector Regression (SVR) [Chun-Hsin *et al.*, 2004] and Neural Networks (NNs) [Lingras *et al.*, 2002]. These methods can capture complex non-linear relations in traffic data but still cannot effectively model spatial-temporal dependencies found in actual traffic.

More recently, deep learning methods have been widely applied in traffic prediction because of their powerful ability to learn high dimensional features and express nonlinear processes. For instance, a Convolutional Neural Network (CNN) captured the spatial dependencies in traffic flow [Ma *et al.*, 2017], while a Recurrent Neural Network (RNN) [van Lint *et al.*, 2005] and Long Short-Term Memory network (LSTM) [Zhao *et al.*, 2017] were employed to extract the temporal dependencies of traffic flows. Approaches integrating CNN and RNN captured both temporal and spatial features [Lv *et al.*, 2018; Yu *et al.*, 2017]. Road networks typically have irregular and non-Euclidean structures, however, standard CNN is restricted to processing regular grid structures (e.g., images and videos). Thus, these methods based on CNN are not appropriate for capturing the complicated spatial correlations underlying a road network.

In order to model traffic conditions on the topological structures of road networks, a graph structure was introduced to represent a road network and Graph Convolution Network (GCN) [Defferrard et al., 2016] was used to capture spatial correlations of traffic data [Yu et al., 2018]. However, the key drawback of GCN is the explicit assumption of an undirected graph, which neglects the directed propagation of traffic flow under the topological structures of road networks, and therefore is unable to express the adjacent upstream and downstream relations in local areas.

Previous studies show that the decomposition-based hybrid methods can effectively improve the prediction accuracy and thus become a popular tool in hydrological and water resources forecasting. Discrete Wavelet Transform (DWT), the most-widely used decomposition method, can decompose the raw data into separate sub-time series and provide a more coherent structure on time-frequency properties of complex time series. This method has also been investigated in recent traffic and passenger forecasting researches. Sun et al. employed DWT to decompose the passenger flow data into different high-frequency and low-frequency series and then predict each frequency series with support vector machines (SVM) [Sun et al., 2015]. Xie et al. utilized DWT to denoise raw traffic volume data and obtained an improved forecasting performance using Kalman filter [Xie et al., 2007]. However, existing deep learning methods rarely exploit these time-frequency properties of traffic speeds when perform forecasting tasks for a whole road network, and therefore cannot extract spatio-temporal patterns from multiple frequency components.

To address these issues, we consider a road network as a directed graph and propose a hybrid prediction method that effectively learns the spatial and temporal dependencies of traffic speeds. The main contributions of this paper can be summarized as follows:

(1) We use discrete wavelet transform (DWT) to decompose raw traffic data into several frequency components, which enables us to exploit time-frequency properties of traffic speeds with different prediction methods.

(2) We employ motifs (local sub-graph structures [Benson et al., 2016]) to define topological structures, especially the downstream and upstream relations in road networks, and integrate these motifs with GCN to construct a motif-based convolution (MGC) filter for directed graphs that captures the spatial correlations of traffic speeds on a road network.

(3) We construct a spatio-temporal network, called Motif-based Graph Convolutional Recurrent Neural Network (Motif-GCRNN) to predict low frequency components of traffic speeds. By incorporating MGC layers and LSTM, Motif-GCRNN is able to extract complex traffic evolution patterns.

(4) Our experimental results on real-world transportation networks demonstrate that the proposed hybrid method outperforms existing baseline methods when forecasting traffic speeds across an entire road network.

## 2 Preliminaries

### 2.1 Problem Formulation

We use an directed graph $G = (V, E, A)$ with $N$ nodes to describe the road network, where nodes $v_i \in V$ denote road segments, edges $(v_i, v_j) \in E$ indicates the directed connectedness from node $v_i$ to node $v_j$ and $A \in R^{N \times N}$ ($A^T \neq A$) represents the adjacency matrix. When there is an edge from node $v_i$ to node $v_j$, $A_{ij} = 1$ otherwise, $A_{ij} = 0$ ($A_{ii} = 0$). We use $x_t^{v_i}$ to represent the average traffic speed on road segment $v_i$ at the time interval $t$ for traffic speed observations. Considering the whole road network, we define a speed vector $X_t = [x_t^{v_0}, x_t^{v_1}, ..., x_t^{v_{N-1}}]$ to represent the speed information of all road segments at the time interval $t$,

Given the historical speed observations until time interval $t$ on road segments in the road network, the task aims to predict traffic speeds at time interval $t+1$, which can be formulated as:

$$X_{t+1} = F\left([X_{t-T}, ..., X_{t-2}, X_{t-1}, X_t], G(V, E, A)\right) \quad (1)$$

where $T$ is the length of historical observations, $G$ is the directed Graph built with the road network, and $F$ is the prediction function that must be learned.

### 2.2 Graph Convolution Network

There are currently two basic approaches to implement graph convolution: the spatial approach and the spectral approach. Spatial approaches rearrange the vertices into the form of a grid, processed by regular convolutional operations. This provides accurate filter localization due to the finite size of the kernel but faces the challenge of matching local neighborhoods [Niepert et al., 2016]. Spectral approaches define the convolution operation via a graph Fourier transform in the spectral domain. Bruna et al. designed graph convolution neural networks based on the spectrum of the graph Laplacian, however their approach has high computational complexity ($O(n^2)$) due to the cost of computing a graph Fourier transform [Bruna et al., 2014]. Defferrard et al. considered the spectral GCN framework with polynomial filters represented by the Chebyshev polynomial (referred to as ChebNet), and can be efficiently computed by applying powers of the graph Laplacian [Defferrard et al., 2016]. The spectral graph convolution can be written as,

$$y = g_\theta(L)x = \sum_{k=0}^{K} \theta_k T_k(\tilde{L})x \quad (2)$$

where $\tilde{L} = 2L / \lambda_{max} - I_n$ is the rescaled Laplacian matrix with maximum Eigenvalue $\lambda_{max}$, the parameter $\theta \in R^K$ is a vector of Chebyshev coefficients, and $T_k(\tilde{L}) \in R^{n \times n}$ is the Chebyshev polynomial of order $k$. $K$ is the number of successive filtering operations or convolutional layers and $K$ localized convolutions effectively exploits the information from the $K-1$-order neighborhood of a node.

## 2.3 Discrete Wavelet Transform

Wavelet transform (WT) is often used to extract information in the analysis of non-stationary data. This transform provides localization properties in both time and frequency domains [Morlet *et al.*, 1982]. In our approach, discrete wavelet transform (DWT) is employed to decompose raw traffic speed data. An effective way to perform DWT is through the Mallat algorithm [Mallat, 1989] that passes data through a series of low-pass and high-pass filters as shown in Equation (3) and (4):

$$A = \sum_{k=-\infty}^{\infty} S[k] \varphi_l [2n-k] \quad (3)$$

$$D = \sum_{k=-\infty}^{\infty} S[k] \varphi_h [2n-k] \quad (4)$$

where $S$ is the original signal, $\varphi_l$ is low-pass filter and $\varphi_h$ is high-pass filter, and $A$ and $D$ are the outputs of low-pass and high-pass filters, called approximation and detail coefficients, respectively.

Figure 1(a) illustrates the process of Mallat algorithm with three-level decompositions. The original time series data $S$ is firstly passed through both low-pass and high-pass filters and sub-sampled by two to obtain the approximation and detail coefficients (i.e., $A_1$ and $D_1$) at the first level. The obtained approximation coefficient $A_1$ is then passed through both filters again to obtain two coefficients, $A_2$ and $D_2$, at the second level. This process is repeated until the specified level has been reached. The last approximation coefficient $A_3$ and all detail coefficients are retained after the decomposition process and wavelet reconstruction is performed on each of these coefficients. Specifically, when reconstructing time series from one coefficient, all coefficients except this one are set as zero values, and fed into the reconstruction algorithm (i.e., inverse discrete wavelet transform, IDWT) to obtain the reconstructed components related to the selected coefficient. As shown in Figure 1(b), the low-frequency component $rA_3$ is reconstructed using the approximation coefficient $A_3$, which exhibits a smooth trend in fluctuation, and high-frequency components with different frequency sub-bands are obtained from reconstruction results of the detail coefficients, which show stochastic change patterns.

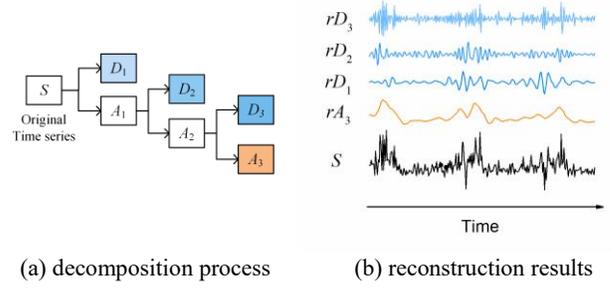

(a) decomposition process  (b) reconstruction results

Figure 1: Discrete Wavelet Transform

## 3 Methodology

### 3.1 Overview

Figure 2 presents the flow chart of our hybrid forecasting method. Original traffic speed data are processed by discrete wavelet transform (DWT) to obtain multiple frequency components for each road segment. A spatial-temporal prediction network – Motif-GCRNN is proposed to capture traffic evolution patterns of the low-frequency components for all road segments, and auto-regressive moving average (ARMA) models are employed to simulate stochastic deterring occurred in the high frequency components. Outputs of different frequency components are added by road segment to obtain the final prediction results.

Figure 3 illustrates the architecture of Motif-GCRNN. As shown in the middle part, we construct motif-based graph convolutional (MGC) layers that capture spatial correlations among road segments and use a recurrent layer with two LSTM that learns temporal dependency of traffic dynamics. At each time interval, the data is a speed vector of all road segments, represented by a simplified notation of a graph to this speed vector as shown in the yellow and blue squares, in Figure 3. We select historical speeds over the past few time intervals and speeds at the same time over the past few days as input to predict the speed at the next time interval. These two parts of input are fed into MGC layers respectively, and then spatial features extracted are reshaped to feed into LSTM for learning short-term and periodic patterns respectively (i.e., *recent trend* and *daily period*). These two types of temporal features are concatenated and fed into a fully connected layer for the prediction outputs.

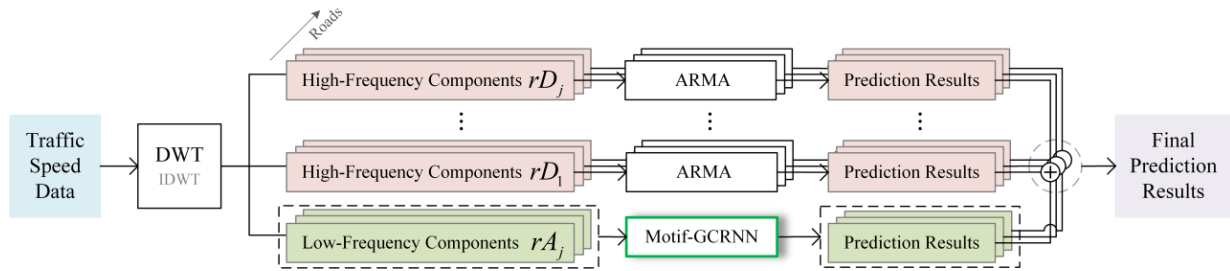

Figure 2: the flow chart of the proposed hybrid prediction method. DWT: Discrete Wavelet Transform; IDWT: Inverse Discrete Wavelet Transform; ARMA: Auto-Regressive Moving Average. Motif-GCRNN: Motif-based Graph Convolutional Recurrent Neural Network.

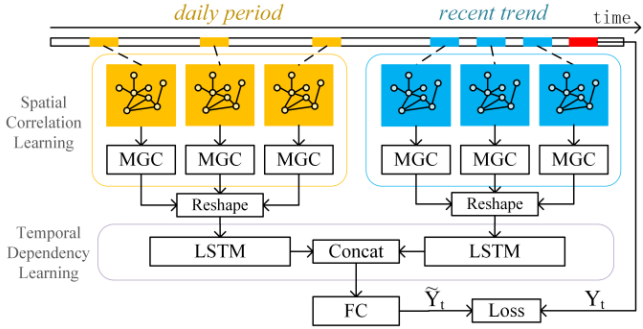

Figure 3: the architecture of Motif-GCRNN. MGC: Motif-based Graph Convolution; LSTM: Long Short-Term Memory; FC: Fully Connected.

### 3.2 Motif-based Graph Convolutional Layer

In a road network, the speed prediction for one road must account for the traffic conditions on adjacent roads. However, existing forecasting methods based on GCN only capture a rough spatial correlation in local areas, neglecting the influence of road direction on traffic flows. Therefore, we introduce a set of special graph structures – motifs into directed graphs to describe local structures in a road network, and we integrate these motifs with spectral GCNs to capture high-order spatial patterns of traffic flows. As traffic speeds in two opposite directions of the same road could differ greatly at the same time interval, they must be predicted separately. Thus in this study, all road segments are assumed to be unidirectional; any road with two lanes going in opposite directions is considered as two unidirectional roads.

**Motifs.** Motifs - small network sub graphs, are considered as fundamental units of complex network and used for exploring certain meaningful connectivity patterns [Benson *et al.*, 2016]. Five motifs with only unidirectional edges among thirteen types of 3-node motifs, as shown in Figure 4, were selected to represent possible local structures in urban road networks. The blue nodes in these motifs are anchor nodes and regarded as the target road segments. More complex road structures can be described by combination of these five motifs.

**Motif-based Adjacency Matrix.** Taking advantage of these sub graphs, we constructed a motif-based adjacent matrix $W_M$

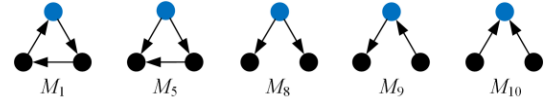

Figure 4: Motifs in Road Networks

that stores the high-order spatial information of a road network. We use an un-weighted directed graph to represent this road network. For each edge $(v_i, v_j) \in E$, let $w_{k,ij}$ denote the number of times the edge $(v_i, v_j)$ participates in $M_k$ ($k \in \{1,5,8,9,10\}$). The motif-based adjacency matrix $W_M$ is defined as:

$$(W_M)_{ij} = \sum_k w_{k,ij} \qquad (5)$$

Each weight in matrix $W_M$ is a statistical value that records the degree of involvement of an edge related to these five motifs for constructing a graph. In a road network, the weight describes the selection probability of vehicles for the connection between two roads, which can be indirectly used to evaluate the importance of this connection in the whole road network. Taking the structure in Figure 5(a) for example, there are three upstream roads (green arrows) and three downstream roads (blue arrows) connecting to the center road with ID of 4. As shown in Figure 5(b), these six edges participate in different motifs according to their context, thus forming different weights for edges as shown in Figure 5(c).

**Motif-based Convolution on Directed Graphs.** Through the weights derived from the adjacency matrix, a local convolution filter can effectively integrate the effects of diverse traffic conditions occurring on upstream and downstream roads to forecast traffic speeds on the central road. Based on the convolution operation as shown in Equation 2, we define the motif Laplacian associated with the adjacency matrix $W_M$ as $\tilde{\Delta} = I - \tilde{D}^{-1/2} W_M \tilde{D}^{-1/2}$ ($\tilde{D}$ is the degree matrix of $W_M$), which acts anisotropically in a preferred direction along structures associated with the motifs. We use this Laplacian to calculate the motif convolution on the directed road graphs $G(V, E, A)$ with signal $X$ and a filter $\Theta$:

$$h_t = \Theta(X_t, G) = \sigma\left(\sum_{k=0}^{K} \theta_k T_k(\tilde{\Delta}) X_t\right) \qquad (6)$$

where $K$ is the size of filters' reception fields, $\sigma$ is the activation function (e.g., ReLU), $\theta \in R^{k \times n}$ is the trainable parameters and $T_k$ is the Chebyshev polynomial of order $k$.

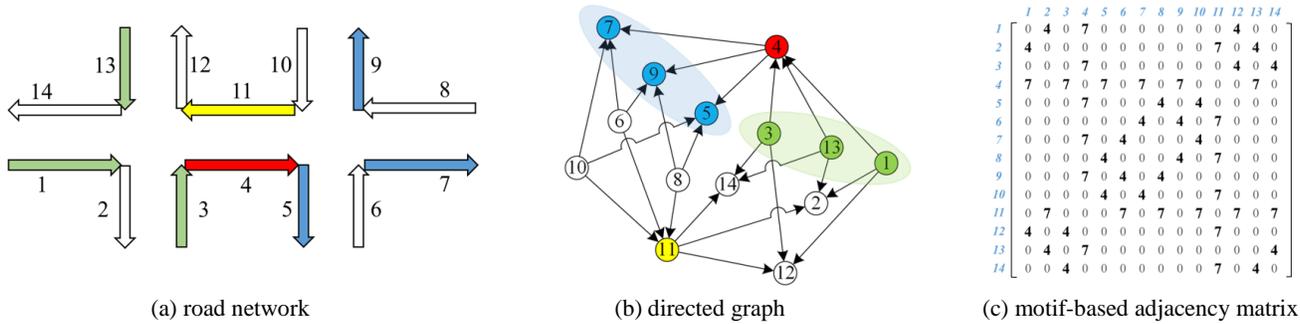

(a) road network  (b) directed graph  (c) motif-based adjacency matrix

Figure 5. Expression of Spatial Relations in Motif-based Adjacency Matrix.

## 3.3 Recurrent Layer

The dynamic trend over a recent time interval and the periodic repetitions at a daily scale in traffic speed data show strong regularity. Taking full advantage of these temporal patterns can help improve prediction performance. As Long Short-Term Memory (LSTM) [Hochreiter *et al.*, 1997] is a powerful deep learning method capable of learning long-term temporal patterns of a time series, we use a recurrent layer with two LSTM to explore the short-term and periodic dependency of traffic speeds, what we term the recent trend and daily period, respectively. Before using the recurrent layer, we reshape the spatial features extracted by MGC layers in the past *m* time intervals and the same time interval of past *n* days into the input form of LSTM separately. In LSTM, the gates structure and the hidden state are unchanged, only the input is replaced by these spatial features. Assume the predicted speed is at the *t* time interval of the *d* day, the recent trend and daily period are defined as:

$$Y_{trend} = LSTM\left(\left[h_{t-m}^{(d)},...,h_{t-2}^{(d)},h_{t-1}^{(d)}\right]\right) \tag{7}$$

$$Y_{period} = LSTM\left(\left[h_t^{(d-n)},...,h_t^{(d-2)},h_t^{(d-1)}\right]\right) \tag{8}$$

To fuse these two types of spatio-temporal features for future traffic prediction, we concatenate them as $Y_C$.

## 3.4 Regression Layer

We use a fully connected layer to learn the comprehensive impact affected by *recent trend* and *daily period*:

$$\tilde{Y}_t = tanh\left(W_{FC} \circ Y_C + b_{FC}\right) \tag{9}$$

where $\circ$ is the element-wise multiplication operator, $W_{FC}$ indicates the learnable parameters, and $b_{FC}$ represents the bias in the fully connected layer.

The Motif-GCRNN can be trained to predict $Y_t$ by minimizing mean squared error between predicted traffic speed vectors and true traffic speed vectors:

$$\ell(\theta) = \sum_t \|Y_t - \tilde{Y}_t\|^2 \tag{10}$$

where $\theta$ are all learnable parameters in Motif-GCRNN.

## 3.5 ARMA Model

Through discrete wavelet transform, the high-frequency components decomposed from original series show a strong random pattern. Since ARMA is one of the most common models used in stationary stochastic process analysis in the time series domain [Ahmed *et al.*, 1979], we employ it to train and predict the high-frequency components of traffic speeds. The outputs will be added together to form the final prediction results. ARMA consists of two polynomials, one for the auto-regression (AR) and the second for the moving average (MA). The calculation process can be represented as.

$$\tilde{Y}_t = c + \varepsilon_t + \sum_{i=1}^p \varphi_i X_{t-i} + \sum_{i=1}^q \lambda_i \varepsilon_{t-i} \tag{11}$$

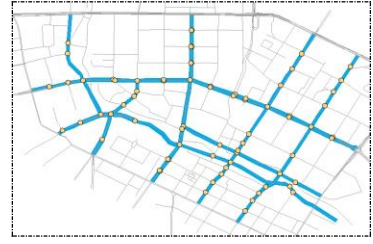

Figure 6: Road network for testing

where *c* is a constant, $\varepsilon$ is the random error and distributed as a Gaussian white noise, $\varphi, \lambda$ are parameters of AR and MA part, *p* and *q* are integers showing the orders of AR and MA part, respectively.

## 4 Experiments

### 4.1 Dataset Description

We verified the proposed approach using a taxi GPS dataset collected from 1st November 2016 to 30th November 2016 (30 days) in Chengdu, China. The highlighted roads in Figure 6 constitute the road network to be predicted, which consists of 156 unidirectional road segments. We calculated the travelling speed between two continuous GPS points of one car as a speed record at a road segment. All speed records during 30 days were aggregated into a 15 min interval, and the traffic speed of each road segment at each time interval was then obtained by averaging the speed records within this time interval for each road segment. In our experiment, data from the first 24 days are used as training data, and the remaining six days as test data.

### 4.2 Experimental Settings

**Baselines**. We compare our approach with the following baseline methods: (1) Auto-Regressive Moving Average (ARMA); (2) Support Vector Regression (SVR); (3) Space-Time Convolutional Neural Network (ST-CNN) [Ma *et al.*, 2017], which uses a speed matrix with x-axis of time and y-axis of roads to rep-resent the spatio-temporal change of traffic speeds; (4) Long Short-Term Memory (LSTM) with two recurrent layers; (5) Graph Convolution Network (GCN) in spectral domain with pooling and fully-connected layers; (6) Graph Convolutional Recurrent Neural Network (GCRNN) [Seo *et al.*, 2018], which integrates GCN and LSTM that exploits both graph spatial and dynamic information to predict structured sequence data..

**Parameter settings.** In our approach, we use one convolutional layer with 32 filters and a max-pooling layer at a size of 2×2. The size of filter reception field *K* varies from one to five for motif-based convolutional layers, the number of recent trend intervals varies from one to eight (15min to 2 hours) and the size of daily period intervals varies from one to eight days for LSTM layers. The filter reception fields *K* was set to three, the number of recent trend intervals to two (half an hour) and daily period intervals were set to seven as default parameters. For DWT, we select level $j = 3$, and used

| Model | MAE | MAPE (%) | RMSE |
|---|---|---|---|
| ARMA | 3.952 | 14.135 | 5.308 |
| SVR | 3.708 | 13.905 | 5.244 |
| ST-CNN | 3.624 | 13.471 | 5.211 |
| GCN | 3.600 | 13.468 | 5.191 |
| LSTM | 3.566 | 13.178 | 5.176 |
| GCRNN | 3.508 | 13.120 | 5.116 |
| Motif-GCRNN | 3.499 | 13.035 | 5.098 |
| **Proposed Method** | **3.287** | **12.112** | **4.700** |

Table 1: Performance comparison of different models

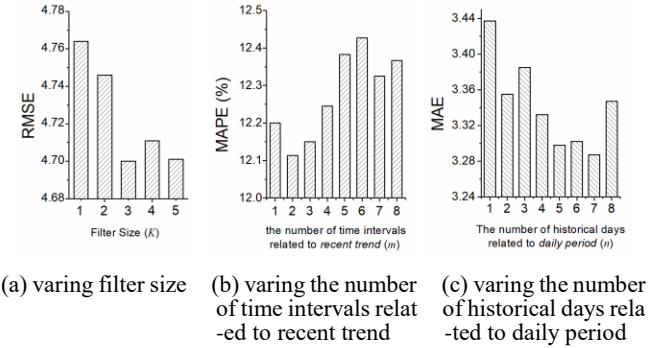

(a) varing filter size  (b) varing the number of time intervals relat-ed to recent trend  (c) varing the number of historical days rela-ted to daily period

Figure 7. Varying different parameters

DB4 as mother wavelet, to obtain three high-frequency components and one low frequency sequence.

We selected Mean Absolute Error (MAE), Mean Absolute Percentage Error (MAPE), and Root Mean Squared Error (RMSE) to evaluate the performance of the tested methods. All neural network based approaches were implemented using Tensorflow, and trained using the SGD optimizer.

### 4.3 Experiment Results

Table 1 illustrates the performance results of our method and other baselines. In the Motif-GCRNN model, the original traffic speed data were directly trained and predicted by Motif-GCRNN without wavelet transform and reconstruction.

We analyzed the results in Table 1 and found that overall, our proposed hybrid prediction method significantly improved performance with the lowest test error among the six other benchmarks tested, according to three metrics. These results verify the effectiveness and superiority of our method. In contrast to Motif-GCRNN, our method achieved higher accuracy for all three metrics, indicating that wavelet transform is valid and the accuracy of speed prediction can be improved by capturing time-frequency properties of traffic speed data. The Motif-GCRNN, GCN, LSTM, ST-CNN and GCRNN comparative results show that Motif-GCRNN has stronger ability to learn spatio-temporal dependency of traffic speeds. Furthermore, Motif-GCRNN outperformed GCRNN, which indicates that motif-based graph convolution extracts more effective high-order spatial features than general graph convolution. We observed that the accuracy of all neural-network based models are higher than ARMA and SVR, which indicates deep learning methods are more suitable for simulating nonlinear spatio-temporal processes than general machine learning methods and traditional statistical methods.

**Effect of different parameters:** Figure 7 shows the effect of different size of filters' reception fields $K$ as well as different number of recent trend intervals and daily period intervals.

As shown in figure 7(a), the RMSE of our method gradually decreases and then fluctuates slightly with an increase of $K$. The best prediction performance occurred when $K = 3$. Intuitively, this may be because a larger K enables the model to capture the broad spatial dependency between the predicted segment and its adjacent road segments. However, the spatial dependency becomes weaker when the involved road segments within the reception field get farther from the predicted segment, indicating that the spatial information of distant road segments with higher adjacent order does not improve the prediction performance.

As shown in figure 7(b), the model achieves the best performance when speed data of previous two time intervals (30 min) are fed into recurrent layers. Then, the MAPE goes up with an increase in historical information. This result indicates that traffic dynamics have a strong short-term temporal dependency, and this dependency is concentrated over a short period in the past. Remote historical information therefore, is less valuable when capturing the temporal patterns of traffic speeds.

As shown in figure 7(c), the prediction accuracy is improved by adding periodic data to the recurrent layer with the increase of historical days. The MAE reaches a relatively low level when the length of *daily period* intervals is more than five. This result shows that traffic conditions at the same time interval during past days can be used to predict the possible traffic conditions in future days. The increase of test error occurred when the length of daily period intervals is eight; this may be due to insufficient training samples with the need for so much historical data input.

## 5 Conclusion and Future Work

In this paper, we propose a novel hybrid method for traffic speed forecasting. The wavelet transform is used to decompose different time-frequency components of raw traffic data. We constructed a spatio-temporal network – Motif-GCRNN to predict the low frequency component and employed ARMA to model high frequency components. In Motif-GCRNN, we learned the road network as a directed graph and defined motif-based traffic graph convolution to capture spatial correlations of traffic speeds among adjacent roads. A recurrent layer with two LSTM is used to learn short-term and periodic information, respectively. The proposed approach outperformed other state-of-the-art methods on a large-scale real-world traffic dataset collected in Chengdu, China.

In future work, we will move forward to explore the performance of different types of mother wavelets in the wavelet transform process, and apply the proposed hybrid method to other spatial-temporal forecasting tasks.